\documentclass{article}
\usepackage{fancyhdr}
\usepackage{spconf,amsmath,graphicx}

\fancypagestyle{firstpage}{
  \fancyhf{}
  \fancyhead[L]{\textit{Proceedings of IC-NIDC 2023}}
  
}

\usepackage{graphicx}
\usepackage{amsmath}
\usepackage{amssymb}
\usepackage{booktabs}
\usepackage{algorithm}
\usepackage{algorithmic}
\usepackage{multirow}

\usepackage{color}

\usepackage{multirow}
\usepackage{bbm}
\usepackage{arydshln}

\usepackage{subfigure}

\newcommand{\cut}[1]{}

\newcommand{\red}[1]{\textcolor{red}{#1}}

\newcommand{\sota}{the state-of-the-art}
\newcommand{\reid}{re-id}

\newcommand{\ours}{Curriculum Person Clustering (CPC)}

\usepackage{url}
\usepackage[capitalize]{cleveref}
\crefname{section}{Sec.}{Secs.}
\Crefname{section}{Section}{Sections}
\Crefname{table}{Table}{Tables}
\crefname{table}{Tab.}{Tabs.}

\title{Unsupervised Long-Term Person Re-Identification with Clothes Change}
\name{Mingkun Li$^{\P, 1}$, Shupeng Cheng$^{\P, 2}$, Peng Xu$^{\dag, 2}$,   Xiantian Zhu$^{\P, 3}$, Chun-Guang Li$^{*, 4}$, Jun Guo$^{*, 4}$}
\address{ 
    $^{1}$ China Academy of Aerospace Science and Innovation,
    $^{2}$Tsinghua University,\\
    $^{3}$University of Surrey,
    $^{4}$Beijing University of Posts and Telecommunications\\
$^\P$\{mingkun.li, chengshupeng4, eddy.zhuxt\}@gmail.com,  \\
$^*$\{lichunguang, guojun\}@bupt.edu.cn, $^\dag$peng$\_$xu@tsinghua.edu.cn
}

\begin{document}
%
\maketitle

\thispagestyle{firstpage}
\begin{abstract}

Most person re-identification methods artificially assume that each person's clothing is stationary in space and time.
Since the average person often changes clothes even within a single day, this condition primarily holds true in situations involving short-term re-identification scenarios.
Some recent studies have investigated re-identification of clothing changes based on supervised learning to reduce this limitation.
In this paper, we remove the necessity for personal identity labels, which makes this new problem dramatically more challenging than conventional unsupervised short-term Re-ID. 
To surmount these obstacles, we introduce a novel approach known as the Curriculum Person Clustering (CPC) method, which exhibits the ability to dynamically modify the clustering criterion based on the clustering confidence in the clustering process.
Experimental results on DeepChange show that  CPC surpasses other unsupervised re-id method and even close to supervised methods.

\end{abstract}
%
\section{Introduction}
\label{sec:intro}
Person re-identification aims to associate the identity of an individual with images acquired from various camera perspectives.  
The majority of re-id methods in use today
\cite{li2019unsupervised, li2020learning}
are based on general scenarios without clothing changes.
This presents a limitation since the majority of individuals alter their attire on a daily basis.
Consequently, their efficacy remains confined to shot-term re-identification task.
This constraint has sparked a burgeoning research interest in long-term person re-id, particularly concerning variations in clothes~\cite{change:6,Change:1}. 
{However, the collection and annotation of personal identity labels is extremely difficult under conditions of unconstrained clothing change.
As shown in Figure~\ref{fig:clothes-change}, due to the diversity of pedestrian appearance, the largest and most realistic clothes change dataset DeepChange \cite{deepchange} was created at great expense.

Recognizing the profound importance of long-term person re-identification and the substantial financial investments required for dataset acquisition, our study centers on addressing the \textit{unsupervised long-term person re-identification} predicament, effectively obviating the necessity for arduous personal identity labeling.
{ Unsupervised long-term is more challenging because of the complexity of different people have similar apperance, while the same person wears different clothes lead to very different appearances.}
Consequently, the current methodologies\cite{Lin:CVPR20,li2021cluster} relying on pseudo-labels would confront formidable dilemmas, ultimately culminating in suboptimal resolutions.

\begin{figure}[!t]
\centering
{\includegraphics[ width=1\columnwidth]{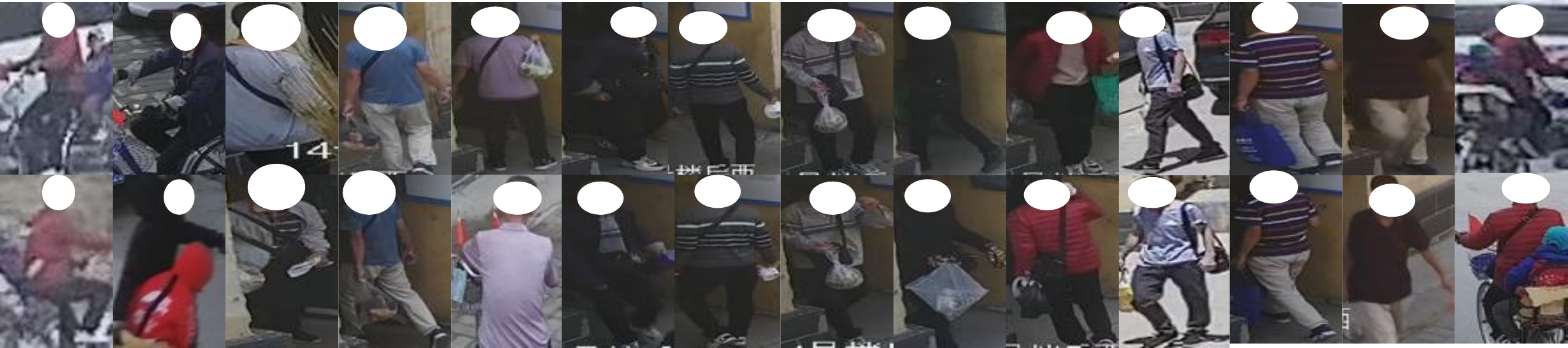}}\\
\caption{Visualizing the inherent challenges of long-term person re-identification.
All the images belong to a single person. It is clear that the difference in appearance between different clothing.
}
\label{fig:clothes-change}
\end{figure}

To address these  obstacles,
we introduce a novel {\bf \ours{}} method.
In order to reduce the accumulation of negative effects of labelling errors throughout the training process, we introduce a pseudo label generating strategy, called curriculum learning clustering.
Specifically, to regulate  the labeling process, we formulate a confidence metric by establishing the correlation between samples within each cluster.
During the training process, only a fraction of the samples will be involved in the training based on the selection of the confidence index, which also means that these samples are currently confident enough to provide correct information for the training of the model.
Also, the confidence index is updated as the training progresses. In this way, the damage caused by incorrect labelled samples for model training can be greatly reduced and the accuracy of the samples used for model training can be improved.
 }

The {\bf contributions} of our CPC as:
{\bf (1)} {To solve unsupervised long-term person re-id challenge, 
we propose the {\bf \ours{}}, which aims to redirect the labelling error that affects propagation in the training process.}
{\bf (2)} Extensive experiments demonstrate that CPC surpasses existing unsupervised methods by a substantial margin, rivaling the performance of fully supervised models on DeepChange, the most largets re-identification benchmark available to date.

\begin{figure}[!t]
\centering
{\includegraphics[ width=1\columnwidth]{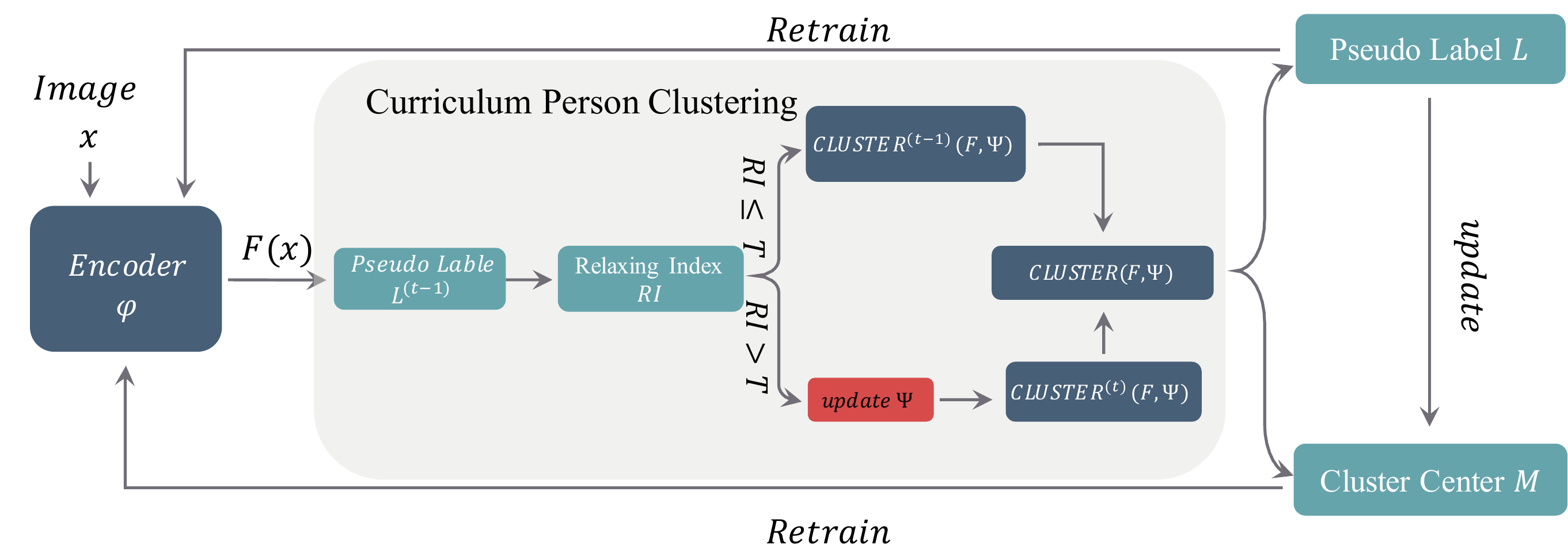}}\\
\caption{The flowchart of \ours{}.
}
\label{fig:pipeline}
\end{figure}

\section{Related Work}
\label{sec:related-work}
\subsection{Long-Term Person Re-id}

Some person re-id methods use fully supervised training to tackle the effects of changing clothes~\cite{Change:1}
These articles essentially try to find other supervisory information in addition to general supervisory information, such as person silhouette, to prompt the model learning cloth independent features.
To make use of the  potential information in the body shape,
Hong~\cite{change:8} extracts pose-specific features and estimates the person's silhouette to effectively utilize the fine-grained features.
{These clothes-changing re-id methods provide inspiring ideas based under supervised training but inevitably need to rely on auxiliary information.}
Henceforth, in this paper, our primary emphasis lies in surmounting the hurdles posed by clothing changing with unsupervised setting.

\subsection{Curriculum Learning}
Curriculum Learning~\cite{CL:4} inspired by the human learning curriculum, that allows the model to have a better understanding of the samples by learning from easy to hard.
Based on the CL, the model can gain better the generalization ability.
Thus, the curriculum learning technique has gained widespread adoption in the train processing of deep learning\cite{CL:2}.
{
{We propose a CL strategy  with unsupervised training for long-term re-id task.}}




\section{Methodology}
\label{sec:method}

We present a novel method known as the \ours{} approach to address the intricate challenge of unsupervised long-term person re-identification. 
Illustrated in Figure~\ref{fig:pipeline}, our CPC framework comprises two modules: (1) the module for acquiring feature representations, and (2) the module dedicated to curriculum-learning-based person clustering.
Within the representation learning module, the outcomes of person clustering through a curriculum learning serve as the supervision information for network training.
As training continues, the enocder's image representation grows in power. In the curriculum person clustering module, we propose an adaptive CL training strategy to automatically optimize the clustering process. This makes the stage-specific clustering select samples based on dynamic criteria.
In CPC, we use ResNet50~\cite{he2016deep} as the encoder network.
We us the penultimate network layer of the model to extract the person image features.
Once we have obtained the clustering results, we use only the clustered samples in the subsequent stage of representation learning.
By leveraging this approach, it becomes possible to mitigate the introduction of errors stemming from pseudo labels, thus reducing their impact to a minimum.


\subsection{Clustering-based Unsupervised Re-id}
\label{sec:representation-learning-and-clustering}

In the absence of supervision information, our focus lies upon training a model using an unlabeled dataset $X = {\{x_n\}}_{n=1}^N$.
In the beginning, we use the network to extract the features ${\bf F} = {\{{\bf f}_n\}^N_{n=1}}$, where ${\bf f}_n = \varphi (x_n, \Theta) \in \mathbb{R}^D$,  where $\varphi(\cdot,\Theta)$ is the feature encoder  with  parameters $\Theta$,  where $D$ is the feature dimension.
Then, we cluster ${\bf F} = {\{{\bf f}_n\}^N_{n=1}}$ by DBSCAN\cite{EsterDBSCAN:AAAI96} to generate pesudo label. 
After clustering, there will be some single samples not fall in any cluster.
Assuming that $Z$ of $N$ samples are clustered together
and we denote the pseudo labels as $L = {\{l_z\}}_{z=1}^Z$, 
and the unclustered $N-Z$ samples will be excluded from this training iteration.
Accordinig to the  pseudo label $L$, we contrast the cluster {center} bank ${\bf M} ={\{{\bf m}_c\}^C_{c=1}} \in \mathbb{R}^{{D\times C}} $, where $C$ is cluster number, ${\bf m }_{c}$ is defined as:
\begin{align}
{\bf m}_{c} = \frac{1}{P_c} \sum_{ l_z = c }  {\bf f}^{l_z},
\end{align}
$P_c$ is the size of cluster $c$, and ${\bf f}^{l_z}$ is the sample features in cluster $c$.

In the training process, we can update the encoder $\bf f$ with parameter set $\Theta$ through the cross-entropy loss function:
    \begin{align}
    \mathcal{L}(\Theta, x_n) = - \ln \frac{\exp({\bf m}_{\omega(x_n)}^\top {\bf f}(x_n, \Theta)/ \tau )}{\sum_{l=1}^C \exp( {\bf m}_{l}^\top {\bf f}(x_n, \Theta) / \tau )},
    \label{eq:repelled-loss}
    \end{align}
where $\omega(x_n)$ is pseudo label index of image $x_n$, $\tau$ is the temperature parameter.  Subsequently, the cluster center bank undergoes an update during the $t$-th iteration in the following manner:
\begin{align}
{{\bf m}_{\omega(x_n)}^{(t)}} \leftarrow \alpha {\bf m}_{\omega(x_n)}^{(t-1)} + (1-\alpha) {\bf f}(x_n),
\label{eq:update-cluster}
\end{align}
where $\alpha$ is the update parameter to control the memory updating. 
Then, we perform the clustering algorithm $CLUSTER(\cdot,\Psi)$ on all $N$ samples to generate new pseudo labels for further training iteration: 
\begin{align}
L = CLUSTER({\bf F},\Psi),
\label{eq:cluster}
\end{align}
where $\Psi$ denotes image clustering algorithm parameters. $L = {\{l_z\}}_{z=1}^Z$ is the pseudo label and is generated by the clustering algorithm, where $Z \leq N$. 
Again, the clustering method will produce a proportion of unclustered images. These samples will not be included in the representation learning in the current iteration.

\subsection{{Curriculum Person Clustering}}
\label{sec:curriculum-person-clustering}
The inherent limitation of the baseline approach lies in its inflexibility to accommodate the diverse range of clothing variations inherent to individual identities. The clustering criterion employed fails to effectively capture these nuanced differences.
To surmount this constraint, we present a novel approach in this section: a dynamic clustering strategy that possesses the ability to dynamically modify the clustering criterion based on the cluster's internal property. 
We use the cluster density index to quantify the difficulty of clustering,  such as clustering confidence, as the measure of the difficulty of learning the curriculum, called the Relaxing Index (RI):
\begin{align}
RI= \frac{1}{Z} \sum_{z=1}^Z  S({\bf f}(x_z), {\bf{m}}_{l_z}),
\label{eq:diffucluty}
\end{align}
the  similarity score $S(\cdot)$ measures the similarity between sample $x$ and its cluster center,
which defined as:
\begin{align}
S(f(x_z),{\bf m}_{l_z})= \frac{\sum_{d=1}^D({\bf f}_{x_z}^{<d>} - \overline{\bf f}_{x_z})({\bf m}_{l_z}^{<d>} - \overline{\bf m}_{l_z})}{\sigma_{{\bf f}_{x_z}} * \sigma_{{\bf m}_{l_z}} },
\label{eq:cov}
\end{align}
where $\overline{\bf f}_{x_z}$ denotes the ${\bf f}_{x_z}$ dimension mean value, and  the ${\bf f}^{<d>}_{x_z}$ is the $d$-th dimension of feature ${\bf f}_{x_z}$.
$\sigma_{{\bf f}_{x_n}}$ and $\sigma_{{\bf m}_{lz}}$ are respectively defined as:
\begin{align}
&\sigma_{ {\bf f}_{x_z}} = {\sum_{d=1}^D \| {\bf f}_{x_z}^{<d>} - \overline{{\bf f}}_{x_z}}\|^2_2,\\
&\sigma_{{\bf m}_{\omega(x_z)}} = {\sum_{d=1}^D \| {\bf m}_{l_z}^{<d>} - \overline{ {\bf m}}_{l_z}\|^{2}_{2}}.
\label{eq:fcov}
\end{align}

{
According to  Eq.\eqref{eq:cov}, a higher cluster density will lead to a larger $RI$,  and a significantly pronounced value of the $RI$ indicates that the current cluster consists of only a solitary item or a small number of clothes. This also implies that the present cluster possesses a promising potential for expansion, enabling it to accommodate and incorporate a greater number of samples.}
We use $RI$ as an indicator parameter to scheduling model training, the \textit{training scheduler} $\delta$ in curriculum learning is defined as:
\begin{align}
\delta = \mathbbm{1} (RI > T),
\label{eq:psi-choice}
\end{align}
where $T$ is a threshold value.

\begin{algorithm}[!t] 
\caption{\ours{}.}
\label{alg:Framwork}
\begin{algorithmic}[1]
\REQUIRE  Long-term Re-ID Dataset $X = {\{x_n\}}_{n=1}^N$.
\STATE {\bf Initialization}: Encoder Model.
\WHILE{epoch $\leq$ 50}
\STATE Extract features ${\bf F}$.
\STATE  Cluster through Eq.~\eqref{eq:cluster} and get $L =\{l_z\}^{Z}_{z=1} $ .
\STATE  Update ${\bf M} $ through Eq.~\eqref{eq:update-cluster};
\STATE Train encoder network through Eq.~\eqref{eq:repelled-loss} and update parameter $\Theta$ 
\STATE Update $RI$ through Eq.~\eqref{eq:diffucluty}.
\STATE Update $CLUSTER(\cdot,\Psi)$ through Eq.~\eqref{eq:psi-update}.
\ENDWHILE
\ENSURE Encoder parameters: $\Theta$.
\end{algorithmic}	
\end{algorithm}

{Then, we introduce an update scheme for $\Psi$ the parameters of person image clustering:}
\begin{align}
\psi^{(t)} = \psi^{(t-1)} + \beta * \delta, \;\; \forall \;\; \psi^{(t-1)} \in \Psi^{(t-1)},
\label{eq:psi-update}
\end{align}
$\beta$ is a hyper-parameter.
Based on the Eq.~\eqref{eq:psi-update}, the model has the capability to gradually expand  the level of clustering , progressing from simpler instances such as identical clothes to more challenging scenarios involving diverse clothing items, by gradually perceiving the latent clothes-independent patterns. Please see our supplementary materials for visualization evaluation.
The summarize of \ours{} are shown in  Algorithm~\ref{alg:Framwork}.

\section{Experiment}
\label{sec:experiment}
\subsection{Experimental Setting}
\label{sec:experimental-setting}

\noindent\textbf{{Datasets}}
\label{p:dataset-and-preprocessing}
Within this section, we assess the performance of CPC on the DeepChange~\cite{deepchange}, which currently stands as the most largest long-term person re-identification dataset. 
It contains $178,407$ images of people with $1,121$ identities from $17$ camera views, collected over $12$ months.

\noindent\textbf{Protocols and metrics}
\label{p:protocols-and-metrics}
To evaluate the performance of the model, we employ two commonly-used metrics for retrieval accuracy: CMC and mAP. 
Nevertheless, in contrast to short-term re-id, long-term re-id requires a more intricate consideration,  the true matches for a given probe image should originate from the same camera but captured at distinct time points, featuring individuals who is adorned in dissimilar clothes.

\noindent\textbf{Competitors}
\label{p:baselines}
{
Certain approaches have shown promising results in addressing unsupervised short-term reid, and these methods often utilize clustering-based models.
To the best of our understanding, no existing methodologies have been specifically designed and tailored for the purpose of addressing unsupervised scenarios on the DeepChange.
In particular, we have selected two commonly used short-term methods as our main competitors: self-paced contrastive learning (SpCL)~\cite{Ge:NIPS20} and cluster contrast (CC)~\cite{tanping:arxiv2020}.}
Furthermore, we have conducted a comparative analysis of our approach with several supervised baseline methods, including MobileNet~\cite{sandler2018mobilenetv2}, OSNet~\cite{zhou2021osnet}, DenseNet~\cite{huang2017densely}), ReIDCaps~\cite{huang2019beyond}, DeiT~\cite{touvron2021training},  BNNeck Re-ID~\cite{luo2020strong} and  Vision Transformer~\cite{dosovitskiy2020image}.

\noindent\textbf{Implementation details}
\label{p:implementation-details}
All experimental procedures were carried out using the PyTorch framework, utilizing the computational power of Nvidia 1080Ti GPU(s). 
For CPC, we employed ResNet50 as the backbone and initialized it ImageNet.
The training procedure have total 50 epochs. The learning rate, initially established at 0.00035, undergoes a decay process whereby it is multiplied by a factor of 0.1 every 20 epochs. Batch size is set to 128.
We additionally introduce several parameters. The temperature parameter $\tau$, assumes a value of 0.05, while the upgrade factor $\alpha$ is set at 0.2. Additionally, the values of $T$, poised at 0.8, and $step$ fixed at 0.01.

\begin{table}[!t]
\begin{center}
\small
\caption{Comparison with \sota{} unsupervised re-identification models on the DeepChange dataset.
$\dagger$ denotes no fine tuning on DeepChange. 
``Clustering Base'' means the use orignial clustering result.
The best results are indicated in \textcolor{red}{red}.}

\resizebox{\columnwidth}{!}{
\setlength{\tabcolsep}{1mm}{
\begin{tabular}{l c c c c}
\hline
{Model} & Rank-1 & Rank-5 & mAP & Backbone\\
\hline\hline

\#1 ResNet50~\cite{he2016deep} $\dagger$        &   15.3&27.3& 02.1 & ResNet50\\
\#2 ViT~\cite{dosovitskiy2020image} $\dagger$      &   11.1&21.7& 01.4  & ViT \\
\hline
\hline

\#3 Clustering Base     &  35.5& 45.1& 09.6  & ResNet50\\
\#4  Clustering Base       &   38.0& 47.6 & 10.5  & ViT\\
\hline
\hline

\#5 SpCL~\cite{Ge:NIPS20}& 32.9& 42.2&  08.6 &  ResNet50 \\

\#6 CC~\cite{tanping:arxiv2020} &37.5 &45.7  &  {10.7}& ResNet50  \\

\#7 SpCL   & 37.2& 46.6 &   10.5 &  ViT\\
\hline
\hline
\#8 \bf CPC   & 45.9 & 54.0  &   \red{14.6} &  ResNet50 \\

\hline

\end{tabular}
}}
\label{table:unsuperivised}
\end{center}
\vspace{-20pt}
\end{table}

\begin{table}[!t]
\setlength{\tabcolsep}{1mm}{
\small
  \caption{
  {Comparison with {\bf supervised} method on DeepChange. }
  }
  \label{table:supervised}
  \centering
  \resizebox{\columnwidth}{!}{
  \begin{tabular}{l   c c c c c}
    \toprule

\multirow{2}{*}{Network/Model} & \multicolumn{4}{c}{Rank} & \multirow{2}{*}{mAP} \\
 \cmidrule(r){2-5}
 &   @1 & @5 & @10 & @20 &  \\
    \midrule
    \midrule
    






\#9 ResNet50~\cite{he2016deep}   &  36.6 & 49.8  &  55.4 & 61.9  & 09.6 \\






\#10 MobileNetv2~\cite{sandler2018mobilenetv2}  &   33.7 & 46.5  & 52.7  & 59.4  & 07.9 \\



\#11
DenseNet121~\cite{huang2017densely}  &   38.2 & 50.2  &  55.9 &  62.4 & 09.1 \\


%



\#12
Inceptionv3~\cite{Szegedy_2016_CVPR}   &    35.0 &  47.7 &  53.9 &  60.6 & 08.8 \\

\hline
\hline

\#13 BNNeck \reid{} ResNet50~\cite{luo2020strong}       &  {47.4} & {59.4}  & {65.1}  & {71.1}  & {12.9} \\

\hline
\hline

\#14 ReIDCaps~\cite{huang2019beyond} (ResNet50)  &  {39.4} & {52.2}  &  {58.8} & {64.9}  & {11.3} \\
   


\hline
\hline

\#15 ViT B16~\cite{dosovitskiy2020image}             &  {49.7} & {61.8}  & {67.3}  & {72.9}  & \red{ 14.9} \\ 


    \hline
\hline






\hline
\hline
\#16 \bf CPC ({\bf Ours, unsupervised}) & 45.9 & 54.0 &58.5 &  63.3 &  {14.6} \\

    \bottomrule
  \end{tabular}}}

\vspace{-15pt}
\end{table}

\subsection{Comparison}
\label{sec:results}
\noindent\textbf{Comparison with unsupervised method}
We retested both methods SPCL~\cite{Ge:NIPS20} and CC~\cite{tanping:arxiv2020} on DeepChange and compared our methods, as shown in Table~\ref{table:unsuperivised}. 
Our approach demonstrates good performance compared to other methods, achieving mAP 14.6\% and a Rank-1  45.9\% on the DeepChange dataset.
{Both SpCL and CC rely on training the model by utilizing high-quality pseudo labels, an efficient strategy for unsupervised methods in the absence of clothes changes\cite{li2022cluster}. However,
it also means that both of these methods heavily depend on the color feature.  Our method can effectively improve this situation without using any auxiliary information.}
{In particular, CPC still has clear advantages over SpCL even with a stronger ViT encoder. This further proves that our method is effective in solving the clothing change challenge.}

\noindent\textbf{Comparison with supervised methods }
In order to further highlight the strengths of our approach in clothing change scenarios, we conducted a comprehensive comparison with numerous supervised competitors.
{As shown in Table~\ref{table:supervised}, even the supervised baselines struggle to effectively address long-term re-identification with low retrieval accuracies. 
This performance proves that the clothes changing  is also very challenging for supervised training. However, our unsupervised model outperforms almost all of the above supervised training methods and achieves a minimal gap to the strongest baseline.}

\begin{table}[!t]
\begin{center}
\small
\caption{{ Ablation study on DeepChange.}
}

\setlength{\tabcolsep}{3mm}{
\begin{tabular}{r c c c c c}
\hline
\multirow{2}{*}{CPC}  &  \multicolumn{4}{c}{Rank} & \multirow{2}{*}{mAP} \\
\cline{2-5}
 & @1 & @5 & @10 & @20 &  \\
\hline
\hline
\#17~~~~~~ 
\textbf{}{\bf $\times$}  & 41.8 &51.1&54.1&  59.2 & 12.4 \\
\#18~~~~~~ 
$\checkmark$ &\bf 45.9 &\bf 54.0 &\bf58.5 &\bf  63.3 &\bf  14.6 \\
\hline
\end{tabular}
}
\label{Tab:ablation}
\end{center}
\vspace{-20pt}
\end{table}

\noindent\textbf{Ablative studies}
To further demonstrate the superiority of CPC, we have conducted a series of ablation studies. The results are shown in Table~\ref{Tab:ablation}. 
The Model (\#17) which excludes the inclusion of the CPC, exhibits performance metrics of only 12.40\% and 41.80\% for mAP and rank-1, respectively. These results fall short when compared to the performance demonstrated by CPC in Model (\#18).
This gap demonstrates the effectiveness of CPC for the clothes change challenge.

{

\vspace{-8pt}
\section{Conclusion}
\label{sec:conclusion}
\vspace{-8pt}

We pay attention to the formidable challenge of unsupervised long-term person re-id in the presence of diverse clothing patterns, because individuals may possess similar attire, while the same person can exhibit a wide range of outfit selections that clearly distinguish them. To surmount this intricate obstacle, we propose a novel method called Curriculum Person Clustering (CPC). This method dynamically adjusts the unsupervised clustering criteria based on the density of the clusters, which can effectively merge images of individuals undergoing clothing changes into the same cohesive clusters. Experiments on the most current and long-term person re-id datasets demonstrate the significant superiority of CPC, even comparable to the supervised re-id models.

\noindent\textbf{Acknowledgment} This paper is supported by the National Natural Science Foundation of China under Grant 61876022.

\newpage
\small
\bibliographystyle{IEEEbib}
\bibliography{renew_template}

\begin{thebibliography}{10}

\bibitem{li2019unsupervised}
Minxian Li, Xiatian Zhu, and Shaogang Gong,
\newblock ``Unsupervised tracklet person re-identification,''
\newblock {\em TPAMI}, 2019.

\bibitem{li2020learning}
Yu-Jhe Li, Zhengyi Luo, Xinshuo Weng, and Kris~M Kitani,
\newblock ``Learning shape representations for clothing variations in person
  re-identification,''
\newblock {\em arXiv preprint arXiv:2003.07340}, 2020.

\bibitem{change:6}
Taiqing Wang, Shaogang Gong, Xiatian Zhu, and Shengjin Wang,
\newblock ``Person re-identification by video ranking,''
\newblock in {\em ECCV}. Springer, 2014.

\bibitem{Change:1}
Qize Yang, Ancong Wu, and Wei-Shi Zheng,
\newblock ``Person re-identification by contour sketch under moderate clothing
  change,''
\newblock {\em TPAMI}, 2019.

\bibitem{deepchange}
Peng Xu and Xiatian Zhu,
\newblock ``Deepchange: A long-term person re-identification benchmark,''
\newblock {\em arXiv preprint arXiv:2105.14685}, 2021.

\bibitem{Lin:CVPR20}
Yutian Lin, Lingxi Xie, Yu~Wu, Chenggang Yan, and Qi~Tian,
\newblock ``Unsupervised person re-identification via softened similarity
  learning,''
\newblock in {\em CVPR}, 2020.

\bibitem{li2021cluster}
Mingkun Li, Chun-Guang Li, and Jun Guo,
\newblock ``Cluster-guided asymmetric contrastive learning for unsupervised
  person re-identification,''
\newblock {\em arXiv preprint arXiv:2106.07846}, 2021.

\bibitem{change:8}
Peixian Hong, Tao Wu, Ancong Wu, Xintong Han, and Wei-Shi Zheng,
\newblock ``Fine-grained shape-appearance mutual learning for cloth-changing
  person re-identification,''
\newblock in {\em CVPR}, 2021.

\bibitem{CL:4}
R{\'e}my Portelas, C{\'e}dric Colas, Lilian Weng, Katja Hofmann, and
  Pierre-Yves Oudeyer,
\newblock ``Automatic curriculum learning for deep rl: A short survey,''
\newblock {\em arXiv preprint arXiv:2003.04664}, 2020.

\bibitem{CL:2}
Guy Hacohen and Daphna Weinshall,
\newblock ``On the power of curriculum learning in training deep networks,''
\newblock in {\em ICML}, 2019.

\bibitem{he2016deep}
Kaiming He, Xiangyu Zhang, Shaoqing Ren, and Jian Sun,
\newblock ``Deep residual learning for image recognition,''
\newblock in {\em CVPR}, 2016.

\bibitem{EsterDBSCAN:AAAI96}
Martin Ester, Hans-Peter Kriegel, J\"{o}rg Sander, and Xiaowei Xu,
\newblock ``A density-based algorithm for discovering clusters in large spatial
  databases with noise,''
\newblock in {\em SIGKDD}, 1996.

\bibitem{Ge:NIPS20}
Yixiao Ge, Feng Zhu, Dapeng Chen, Rui Zhao, and Hongsheng Li,
\newblock ``Self-paced contrastive learning with hybrid memory for domain
  adaptive object re-id,''
\newblock in {\em NeurIPS}, 2020.

\bibitem{tanping:arxiv2020}
Zuozhuo Dai, Guangyuan Wang, Siyu Zhu, Weihao Yuan, and Ping Tan,
\newblock ``Cluster contrast for unsupervised person re-identification,''
\newblock {\em arXiv preprint arXiv:2103.11568}, 2021.

\bibitem{sandler2018mobilenetv2}
Mark Sandler, Andrew Howard, Menglong Zhu, Andrey Zhmoginov, and Liang-Chieh
  Chen,
\newblock ``Mobilenetv2: Inverted residuals and linear bottlenecks,''
\newblock in {\em CVPR}, 2018.

\bibitem{zhou2021osnet}
Kaiyang Zhou, Yongxin Yang, Andrea Cavallaro, and Tao Xiang,
\newblock ``Learning generalisable omni-scale representations for person
  re-identification,''
\newblock {\em TPAMI}, 2021.

\bibitem{huang2017densely}
Gao Huang, Zhuang Liu, Laurens Van Der~Maaten, and Kilian~Q Weinberger,
\newblock ``Densely connected convolutional networks,''
\newblock in {\em CVPR}, 2017.

\bibitem{huang2019beyond}
Yan Huang, Jingsong Xu, Qiang Wu, Yi~Zhong, Peng Zhang, and Zhaoxiang Zhang,
\newblock ``Beyond scalar neuron: Adopting vector-neuron capsules for long-term
  person re-identification,''
\newblock {\em TCSVT}, 2019.

\bibitem{touvron2021training}
Hugo Touvron, Matthieu Cord, Matthijs Douze, Francisco Massa, Alexandre
  Sablayrolles, and Herv{\'e} J{\'e}gou,
\newblock ``Training data-efficient image transformers \& distillation through
  attention,''
\newblock in {\em ICML}, 2021.

\bibitem{luo2020strong}
Hao Luo, Wei Jiang, Youzhi Gu, Fuxu Liu, Xingyu Liao, Shenqi Lai, and Jianyang
  Gu,
\newblock ``A strong baseline and batch normalization neck for deep person
  re-identification,''
\newblock {\em TMM}, 2020.

\bibitem{dosovitskiy2020image}
Alexey Dosovitskiy, Lucas Beyer, Alexander Kolesnikov, Dirk Weissenborn,
  Xiaohua Zhai, Thomas Unterthiner, Mostafa Dehghani, Matthias Minderer, Georg
  Heigold, Sylvain Gelly, et~al.,
\newblock ``An image is worth 16x16 words: Transformers for image recognition
  at scale,''
\newblock {\em arXiv preprint arXiv:2010.11929}, 2020.

\bibitem{Szegedy_2016_CVPR}
Christian Szegedy, Vincent Vanhoucke, Sergey Ioffe, Jon Shlens, and Zbigniew
  Wojna,
\newblock ``Rethinking the inception architecture for computer vision,''
\newblock in {\em CVPR}, 2016.

\bibitem{li2022cluster}
Mingkun Li, Chun-Guang Li, and Jun Guo,
\newblock ``Cluster-guided asymmetric contrastive learning for unsupervised
  person re-identification,''
\newblock {\em IEEE Transactions on Image Processing}, vol. 31, pp. 3606--3617,
  2022.

\end{thebibliography}

\end{document}